\documentclass[12pt]{article}

\usepackage[utf8]{inputenc}
\usepackage{amsmath, amssymb}
\usepackage{graphicx}
\usepackage{authblk}
\usepackage[a4paper, margin=1in]{geometry}
\usepackage[superscript]{cite}
\usepackage{setspace}
\usepackage{lineno}
\usepackage[colorlinks=true,
            linkcolor=red,
            urlcolor=red,
            citecolor=blue]{hyperref}
\usepackage{multirow}
\usepackage{url}
\usepackage{xcolor}
\begin{document}

% Title
{\noindent\Large\textbf{NeuroRAD-FM: A Foundation Model for Neuro-Oncology with Distributionally Robust Training
}}

\vspace{1em}

% Authors and affiliations
\noindent{\large Moinak Bhattacharya\textsuperscript{a}, Angelica P. Kurtz\textsuperscript{b}, Fabio M. Iwamoto\textsuperscript{c}, Prateek Prasanna*\textsuperscript{a}, Gagandeep Singh*\textsuperscript{b}}

\vspace{1em}

\textsuperscript{a}Department of Biomedical Informatics, Stony Brook University, Stony Brook, NY, USA

\textsuperscript{b}Department of Radiology, Columbia University Irving Medical Center, NY, USA

\textsuperscript{c}Department of Neuro-Oncology, Columbia University Irving Medical Center, NY, USA

% \textsuperscript{1}Present address: Department of Imaging Science, Another University, City, Country

\vspace{2em}

% Corresponding author
\noindent\textbf{Corresponding Author:} \\
Prateek Prasanna, PhD \\
Stony Brook University \\

\noindent Email: prateek.prasanna@stonybrook.edu

\newpage

\begin{abstract}
\noindent\textbf{Background.} Neuro-oncology presents unique challenges for machine learning due to heterogeneous data distributions
%heterogeneity of MRI acquisition protocols 
and the biological complexity of brain tumors. As a result, generalizing foundation models (FMs) across diverse cohorts remains difficult. A further limitation is the poor performance of existing FMs in predicting uncommon molecular markers—an area critical for treatment response assessment and risk stratification. To address these gaps, our proposed FM incorporates a distributionally robust loss function, enabling accurate estimation of tumor-specific phenotypes while maintaining strong generalization across institutions.
%Traditional radiomics pipelines rely on tumor segmentation, making them sensitive to inter-site variability and prone to performance degradation under domain shift. Similarly, supervised convolutional neural networks (CNNs) require extensive annotated datasets, which are costly and often infeasible at scale. To address these limitations, we propose a segmentation-free, multi-sequence MRI foundation model leveraging T1, T1-contrast-enhanced, T2, and FLAIR sequences. Our framework is designed to capture tumor-specific phenotypes while maintaining robustness across institutions, and we systematically investigate whether imbalance-aware training strategies enhance model transferability to downstream molecular and survival prediction tasks.

\noindent\textbf{Materials and methods.} We pretrain self-supervised backbones (BYOL, DINO, MAE, and MoCo) on multi-institutional brain tumor MRI and incorporate distributionally robust optimization (+DRO) to mitigate site and class imbalance. Downstream tasks include (i) molecular classification (MGMT, IDH1, 1p/19q, EGFR; uncommon alterations ATRX, TP53, CDKN2A/2B, TERT; continuous markers Ki-67 and TP53) and (ii) overall survival in IDH1 wild-type glioblastoma at UCSF, UPenn, and CUIMC. Site invariance is assessed via PCA-silhouette and t-SNE; interpretability is examined with Grad-CAM.

\noindent\textbf{Results.} Our method (+DRO) consistently improved molecular prediction on the combined test-set and reduced difference in feature embeddings across different cohorts.
%site clustering in embedding space. 
At CUIMC (single-site analysis of six targets), mean balanced accuracy increased from 0.744 to 0.785 and AUC from 0.656 to 0.676; under-represented endpoints saw the largest gains (e.g., CDKN2A/2B balanced accuracy 0.86 to 0.92, AUC 0.73 to 0.92; ATRX AUC 0.69 to 0.82; Ki-67 balanced accuracy 0.60 to 0.69). For survival, c-index improved with +DRO at all sites: CUIMC 0.592 to 0.597, UPenn 0.647 to 0.672, UCSF 0.600 to 0.627.
%; +DRO achieved 0.628$\pm$0.058 (MAE at CUIMC), 0.672$\pm$0.068 (MAE at UPenn), and 0.690$\pm$0.065 (BYOL at UCSF). 
Grad-CAM focused on tumor and peri-tumoral regions across markers proving the interpretability of the proposed method.

\noindent\textbf{Conclusion.} A neuro-oncology–specific foundation model coupled with DRO yields more site-invariant representations, improves molecular prediction—including for uncommon markers—and enhances survival discrimination across independent cohorts. These findings underscore the need for prospective validation with calibration and utility analyses, and for integration of longitudinal and interventional signals to advance precision neuro-oncology into clinical practice.

\end{abstract}

\clearpage
\section*{Introduction}
Brain tumors represent a significant global health challenge, contributing substantially to cancer-related morbidity and mortality. According to recent estimates from the Global Cancer Observatory (GLOBOCAN) and the Central Brain Tumor Registry of the United States (CBTRUS), primary brain and central nervous system (CNS) tumors account for approximately 1.4\% of all cancers worldwide but are responsible for a disproportionately high burden of disease due to their aggressive nature and poor prognosis~\cite{price2024cbtrus,GLOBOCAN2022}. Specifically, malignant brain tumors such as glioblastoma (GBM) exhibit a dismal five-year survival rate of less than 7\%, despite advances in surgical techniques, chemotherapy, and radiotherapy~\cite{stupp2005radiotherapy}. Even lower-grade gliomas, which initially present a more indolent course, eventually progress to higher grades, severely impacting long-term outcomes~\cite{yan2009idh1,eckel2015glioma}. Accurate diagnosis, prognostication, and molecular characterization of brain tumors are therefore critical for guiding clinical decision-making, improving patient outcomes, and facilitating personalized treatment strategies. 

Magnetic resonance imaging (MRI) remains the cornerstone of brain tumor diagnosis, treatment planning, and longitudinal monitoring. MRI provides superior soft tissue contrast and enables detailed visualization of tumor morphology, peritumoral edema, necrosis, and infiltration patterns. It also allows non-invasive assessment of tumor heterogeneity across multiple MRI sequences, including T1-weighted (T1), T1-weighted contrast-enhanced (T1ce), T2-weighted (T2), and Fluid-Attenuated Inversion Recovery (FLAIR). These sequences collectively capture diverse tissue characteristics, including blood-brain barrier disruption, fluid content, and tumor cellularity. Consequently, MRI plays an indispensable role in both the clinical workflow and research aimed at understanding brain tumor biology~\cite{ellingson2017modified}.

Over the past decade, quantitative imaging methods such as radiomics have become popular tools for characterizing brain tumors from MRI. Radiomics involves extracting a high-dimensional set of handcrafted features that quantify tumor shape, texture, intensity, and wavelet transformations from segmented tumor regions. These features are typically derived from anatomically defined subregions of the tumor, including the enhancing tumor core, necrotic core, peritumoral edema, and occasionally, peri-tumoral brain tissue. Numerous studies have demonstrated the potential of radiomics features in predicting clinical endpoints such as patient survival, molecular biomarkers like IDH mutation and MGMT promoter methylation, and treatment response~\cite{menze2014multimodal,bakas2017advancing}. Despite these successes, radiomics approaches suffer from several critical limitations. The extraction of radiomics features is highly dependent on accurate tumor segmentation, which is both labor-intensive and prone to interobserver variability. Manual segmentation is not scalable for large datasets, and even semi-automated methods often require substantial expert correction. Moreover, radiomics features are sensitive to imaging acquisition parameters, scanner types, and institutional protocols, limiting their generalizability across multi-center datasets. Scanner-related variability can introduce significant bias, making radiomics models brittle when applied outside the domain in which they were developed. Furthermore, handcrafted features inherently encode assumptions about what constitutes relevant image information, potentially omitting latent and complex patterns that are visually imperceptible yet biologically significant. This rigid feature engineering pipeline stands in contrast to the growing success of deep learning, which automatically learns hierarchical features directly from data without manual intervention. Deep learning models, particularly convolutional neural networks and transformer-based architectures, have demonstrated remarkable success in a wide range of medical imaging tasks, including tumor segmentation, image registration, classification, and disease~\cite{isensee2021nnu}.

Recently, foundation models have revolutionized machine learning, extending rapidly from natural language processing to computer vision and biomedical imaging~\cite{moor2023foundation}. Foundation models are large-scale pre-trained models that learn generalizable representations from vast, diverse, and often unlabeled datasets. These models are designed to be fine-tuned or adapted to a wide array of downstream tasks with minimal additional training. Compared to traditional supervised learning, foundation models require far fewer task-specific annotations and exhibit superior performance, generalization, and robustness across diverse tasks and datasets. 

Several medical imaging foundation models have recently been developed to enable general-purpose tasks without requiring large annotated datasets. Early advances such as MedSAM extended the vision of the Segment Anything Model to clinical domains, demonstrating that large-scale pretraining could generalize across diverse segmentation problems~\cite{ma2024segment}. Parallel efforts in vision–language learning, exemplified by MedBLIP~\cite{chen2024medblip}, RETFound for retinal disease detection~\cite{zhou2023foundation}, and REFERS for chest radiographs~\cite{zhou2022generalized}, illustrated how multimodal supervision can transfer effectively to diagnostic applications. Radiology-specific initiatives including RadImageNet~\cite{mei2022radimagenet} and the recently introduced RadFM~\cite{wu2025towards} provided large-scale backbones tailored to medical image interpretation, while pathology-driven models such as Prov-GigaPath~\cite{xu2024whole} highlighted the scalability of foundation models across modalities. Another recent oncology-focused FM focused on biomarker discovery~\cite{pai2024foundation}. Together, these developments mark a shift toward generalist medical AI~\cite{moor2023foundation}. In neuroimaging, emerging MRI-based foundation models have been applied to sequence type classification~\cite{tak2024foundation}, brain age prediction~\cite{cole2017predicting}, and tissue segmentation~\cite{sun2025foundation}, showing utility in neurodevelopmental and neurodegenerative studies. However, these efforts primarily address healthy brain anatomy, leaving oncologic neuroimaging largely unexplored~\cite{paschali2025foundation}.

Despite the rapid progress in general neuroimaging foundation models, there remains a significant gap in developing models specifically designed for neuro-oncology applications. A major limitation in training such foundation models is the scarcity of large, high-quality neuro-oncology datasets~\cite{bakas2017advancing}. Brain tumors exhibit complex, heterogeneous imaging patterns distinct from healthy brain tissue, including necrosis, enhancement, infiltration, and mass effect. The challenges in brain tumor imaging are fundamentally different from those in healthy brain MRI analysis. Therefore, a foundation model trained on healthy brain structures is unlikely to capture the pathological signatures necessary for tasks like molecular biomarker prediction, tumor grading, or survival estimation. Neuro-oncology demands specialized solutions for key downstream tasks, including prediction of molecular markers (e.g., MGMT promoter methylation, IDH1 mutation, 1p/19q codeletion, EGFR amplification) that are now integral to the WHO brain tumor classification and directly inform therapeutic decisions~\cite{kickingereder2016radiogenomics}. Equally important are prognostic applications, such as estimating overall survival, time to progression, and treatment response, which underpin personalized care. Traditional machine learning approaches—whether handcrafted radiomics pipelines or conventional CNN-based models—have shown only modest performance in these settings, with limited generalizability across institutions~\cite{chang2018residual}.

In this study, we propose \textbf{the first foundation model tailored specifically for neuro-oncological applications} using multi-sequence brain MRI. To the best of our knowledge, this is the first large-scale effort that leverages pretraining on heterogeneous brain tumor datasets to learn tumor-specific representations that generalize across a range of oncologic downstream tasks. Existing foundation models are not optimized to handle out-of-distribution data, often resulting in limited robustness and poor generalization to unseen clinical scenarios. Our model is designed to overcome the limitations of prior approaches in three key ways. First, by eliminating the need for manual tumor segmentation, our approach enables feature extraction and representation learning from large-scale multi-institutional datasets. Second, our model learns robust features that are transferable across diverse downstream tasks, including molecular biomarker prediction (MGMT, IDH1, EGFR), uncommon marker prediction (CDKN2A/2B, TERT, ATRX, TP53, Ki67, etc.), and overall survival prediction. Third, by leveraging self-supervised learning, our model minimizes reliance on large labeled datasets—such as genomics annotations (e.g, IDH, MGMT, EGFR)—and requires only minimal supervision during fine-tuning to achieve state-of-the-art performance. By bridging the gap between general MRI foundation models and the unique demands of neuro-oncology, our approach represents a significant step toward precision medicine for brain tumor patients. This work not only advances the technical capabilities of medical imaging AI but also has the potential to directly impact clinical workflows by providing reliable, scalable, and automated tools for diagnosis, prognosis, and treatment planning in neuro-oncology.\\
The primary contributions of the proposed work are:
\begin{itemize}
    \item We propose \textbf{NeuroRAD-FM}, a novel \textbf{R}obust-\textbf{A}cross-\textbf{D}istribution Foundation Model trained on a large cohort of brain MRI images for neuro-oncology.
    \item We propose a novel distributionally robust optimization method for training the foundation model to ensure its generalizability across different cohort distributions.
    \item We demonstrate our result on 2 public cohorts and 1 in-house cohort for different uncommon mutation predictions.
\end{itemize}

\section*{Materials and methods}
\begin{figure}
    \centering
    \includegraphics[width=1\linewidth]{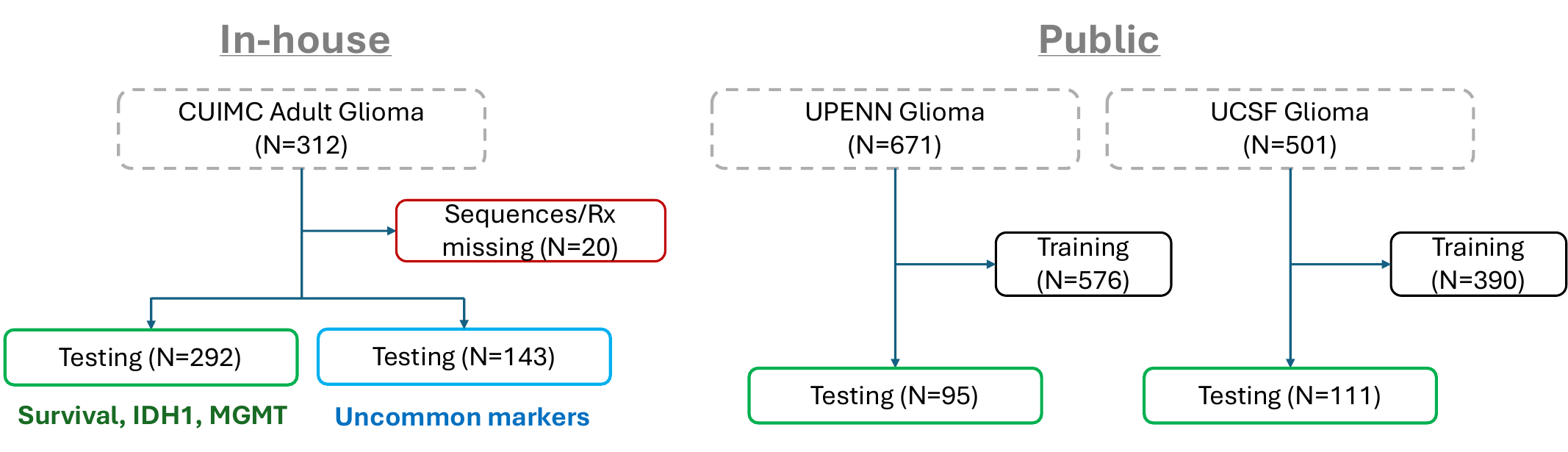}
    \caption{Inclusion-exclusion criteria}
    \label{fig:figure1}
\end{figure}
\textbf{Study design and overview:} This study aims to develop a brain tumor-specific foundation model using multi-sequence MRI for downstream neuro-oncology tasks such as molecular biomarker prediction and survival analysis. We leveraged multi-institutional datasets, including the BraTS challenges and an independent institutional cohort, comprising pre-operative MRI scans with T1, T1ce, T2, and FLAIR sequences. The model was pretrained using a self-supervised masked autoencoding strategy on full 3D MRI volumes without requiring tumor segmentations, enabling it to capture tumor-specific and anatomical features. The pretrained model was subsequently fine-tuned for tasks including MGMT, IDH1, 1p/19q, and EGFR status prediction, glioma grading, and overall survival analysis. MRI data were standardized with skull stripping, bias correction, registration, resampling, and intensity normalization to ensure consistency across institutions. Model performance was evaluated using AUC for classification tasks and c-index for survival prediction, with external validation performed on an independent test cohort to assess generalizability.\\
\textbf{Study Population:} The training dataset comprised MRI scans from 7,000 adult glioma patients aggregated from the BraTS~\cite{maleki2025analysis}, MU-Glioma-Post~\cite{de20242024}, and LUMEIRE~\cite{suter2022lumiere} datasets, representing diverse multi-institutional sources. For testing, we utilized an independent multi-institutional cohort including patients from Columbia University Irving Medical Center (CUIMC), the University of Pennsylvania (UPENN)~\cite{bakas2022university}, and the University of California, San Francisco (UCSF)~\cite{calabrese2022university}. All patients in the testing cohort are diagnosed with adult gliomas and have comprehensive molecular and clinical information, including MGMT promoter methylation status, EGFR amplification, IDH1 mutation, 1p/19q codeletion, and overall survival outcomes. This diverse test cohort was used to rigorously evaluate the model's performance on clinically relevant downstream tasks. Inclusion and Exclusion criteria are shown in Figure~\ref{fig:figure1}.\\
\begin{figure}
    \centering
    \includegraphics[width=1\linewidth]{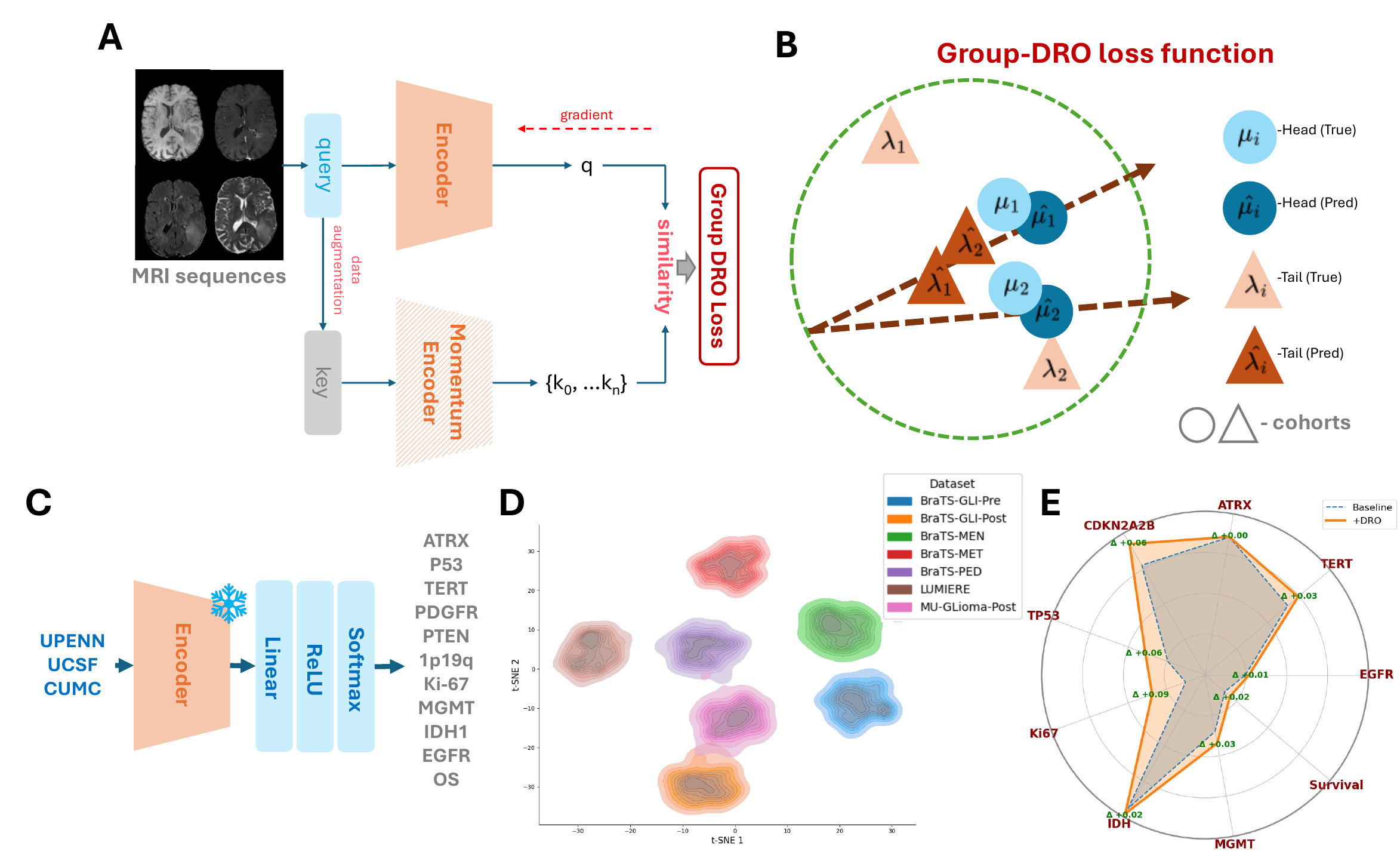}
    \caption{A. \textbf{Pipeline:} Contrastive learning framework for training the foundation model in a self-supervised manner, B. Group-DRO loss, C. Finetuning the the pre-trained encoder for downstream tasks, D. Visualization of distribution of different datasets, E. Summary of the results for Group-DRO compared to baseline approaches.}
    \label{fig:figure2}
\end{figure}
\noindent\textbf{Foundation model training:} Foundation model training was conducted using a large, heterogeneous dataset comprising multiple cohorts and disease types. To address distributional shifts across institutions and disease subtypes, we employed a Group Distributionally Robust Optimization (Group-DRO) loss, which encourages the model to learn representations that are robust across different subgroups. The model architecture is based on a 3D ResNet50 encoder, trained using a self-supervised learning framework (shown in Figure~\ref{fig:figure2}A). As baseline comparisons, we also implemented standard self-supervised learning methods, including MoCo~\cite{he2020momentum}, DINO~\cite{caron2021emerging}, MAE~\cite{he2022masked} and BYOL~\cite{grill2020bootstrap}, using identical training protocols. We hypothesize that incorporating Group-DRO loss (Figure~\ref{fig:figure2}B) improves the generalizability of the learned representations, leading to superior performance on downstream neuro-oncology tasks. All models were trained for 100 epochs with a learning rate of 1e-5 on an NVIDIA A6000 GPU (52 GB VRAM).\\
\textbf{Molecular subtyping:} For molecular subtyping tasks, feature embeddings extracted from the pretrained foundation model were used as inputs to a linear classifier (shown in Figure~\ref{fig:figure2}C). This classifier was trained to predict key molecular markers, including MGMT promoter methylation, IDH1 mutation, 1p/19q codeletion, and EGFR amplification. By leveraging the rich imaging representations learned during self-supervised pretraining, the linear classifier enables efficient and accurate prediction of molecular subtypes directly from MRI features.\\
\textbf{Survival analysis:} For survival analysis, features were extracted from the pretrained foundation model by passing the multi-sequence MRI scans through the model's encoder. The resulting feature embeddings were then used as input to a Cox proportional hazards (Cox-PH) model to predict overall survival. This approach enables the model to capture complex imaging-derived representations while leveraging the Cox-PH model's capability to handle time-to-event data with censored observations.\\
\textbf{Statistical analysis:} Model performance was evaluated across multiple downstream tasks, including molecular biomarker prediction and overall survival analysis. For binary classification tasks, performance was assessed using the area under the receiver operating characteristic curve (AUC). For survival prediction, the concordance index (c-index) was used to measure the agreement between predicted risk scores and actual survival outcomes. Kaplan-Meier survival curves were generated by stratifying patients into high- and low-risk groups based on the median predicted risk score, and differences were tested using the log-rank test. Statistical significance between model performances was assessed using paired t-tests or Wilcoxon signed-rank tests based on the distribution normality. 
%Feature importance for downstream tasks was analyzed using SHAP (SHapley Additive exPlanations) values derived from the fine-tuned models. 
All statistical analyses were conducted using Python (v3.9) with Scikit-learn, Lifelines, and SciPy libraries, with a significance threshold set at p $<$ 0.05.
\section*{Results}
% \textbf{Patient Characteristics.}\\
\begin{figure}[h]
    \centering
    \includegraphics[width=1\linewidth]{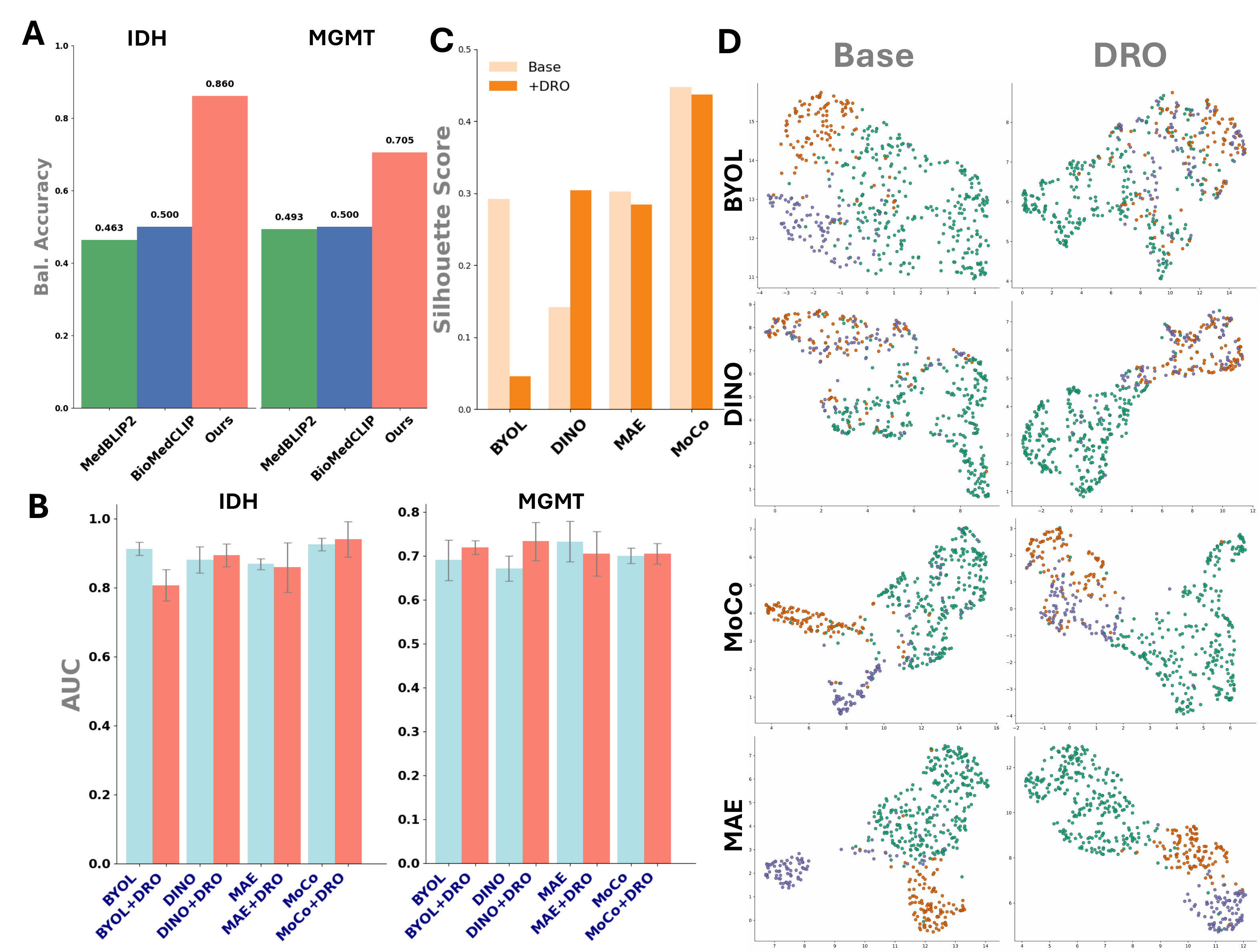}
    \caption{A. Comparison of our (Base+DRO) method with existing FM baselines, B. Comparison of different base SSL methods and +DRO for IDH and MGMT prediction on a combined cohort, C. Silhoutte score for the different base SSL methods and +DRO for robustness estimation, D. t-SNE plots for different SSL methods.}
    \label{fig:figure3}
\end{figure}
\begin{figure}
    \centering
    \includegraphics[width=0.85\linewidth]{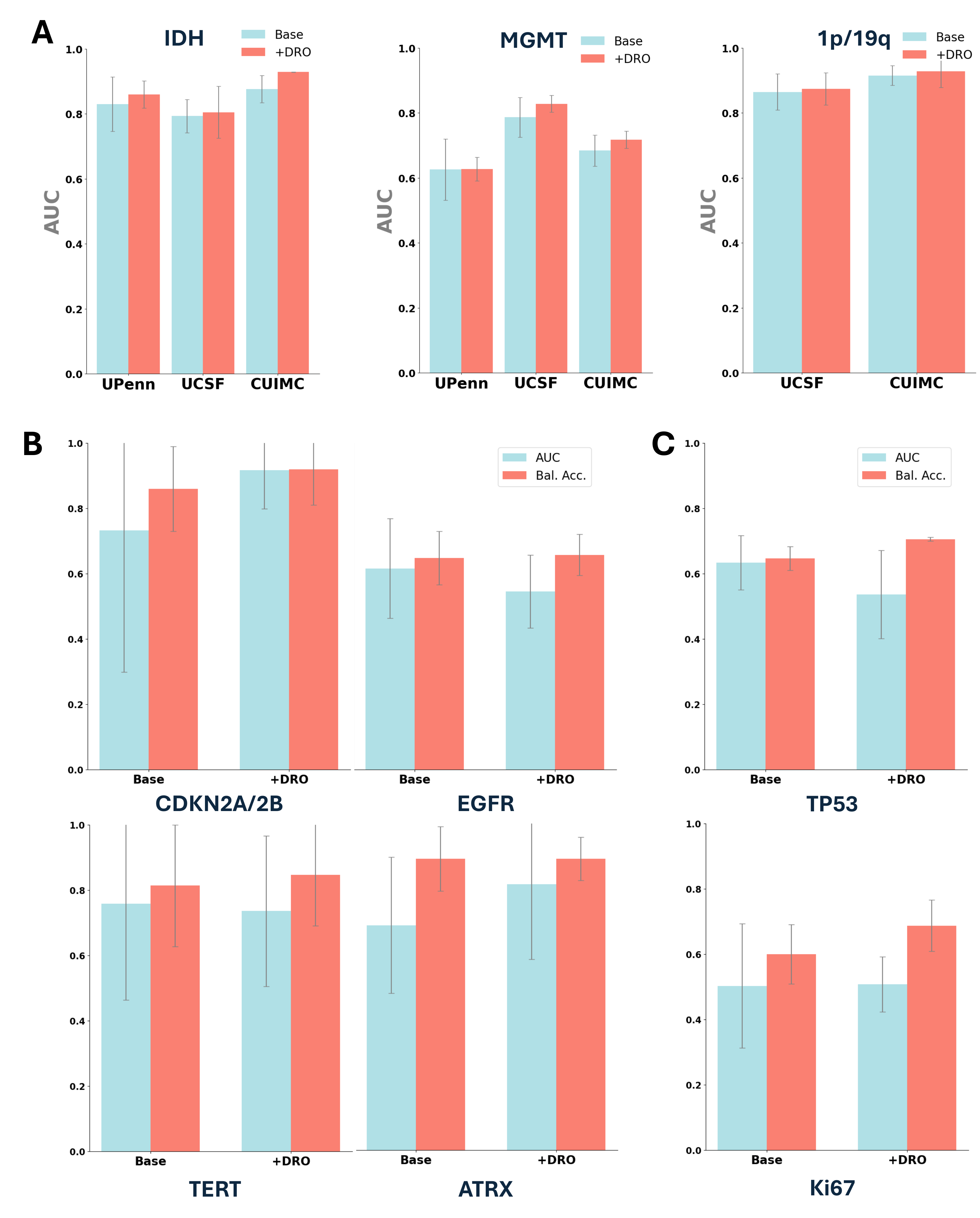}
    \caption{A. IDH, MGMT, and 1p/19q marker prediction for Upenn, UCSF, and CUIMC cohorts, B. Uncommon markers like CDKN2A/2B, EGFR, TERT, ATRX prediction on CUIMC cohort, C. TP53 and Ki67 marker prediction on CUIMC cohort.}
    \label{fig:figure4}
\end{figure}

\textbf{Multi-Genomic Marker Prediction:} We evaluated the effect of incorporating distributionally robust optimization (DRO) into standard self-supervised foundation models (BYOL, DINO, MAE, and MoCo) for predicting key molecular alterations from MRI-derived embeddings. In Figure~\ref{fig:figure3}A, we compare our proposed method with existing methods like MedBLIP2~\cite{chen2024medblip} and BioMedCLIP~\cite{zhang2023biomedclip} on IDH1 and MGMT status prediction. On the combined multi-center test set, DRO-enhanced models consistently outperformed their base counterparts in the classification of MGMT promoter methylation and IDH1 mutation status (Figure~\ref{fig:figure3}B). Cohort separability analysis using PCA-based silhouette scores (Figure~\ref{fig:figure3}C) and t-SNE projections (Figure~\ref{fig:figure3}D) demonstrated that +DRO embeddings exhibited lower site-specific clustering compared with base models, indicating improved site invariance. Center-specific performance analyses further confirmed the robustness of +DRO across datasets: MGMT (UCSF, UPenn, CUIMC), IDH1 (UCSF, UPenn, CUIMC), 1p/19q codeletion (UCSF, CUIMC) (Figure~\ref{fig:figure4}A).\\
% , uncommon mutations including ATRX, EGFR, CDKN2A/2B, and TERT (binary classification; Figure~\ref{fig:figure4}B), and continuous molecular markers Ki-67 and TP53 (Figure~\ref{fig:figure4}C). These findings indicate that DRO not only enhances predictive accuracy but also improves the biological relevance of learned representations by mitigating site-specific biases.\\
\noindent\textbf{Uncommon Markers Prediction through Robust Reweighting:} We evaluated imbalance-aware training (distributionally robust optimization, +DRO) at a single site (CUIMC) for six molecular targets, reporting AUC and balanced accuracy. In Figure~\ref{fig:figure4}B, uncommon mutations including ATRX, EGFR, CDKN2A/2B, and TERT (binary classification are reported and in Figure~\ref{fig:figure4}C, continuous molecular markers Ki-67 and TP53 are reported. On average, balanced accuracy increased from 0.744 to 0.785 and AUC from 0.656 to 0.676. Balanced accuracy improved for all six markers. The largest gains were on under-represented endpoints: CDKN2A/2B improved from 0.86 to 0.92 (AUC 0.73 to 0.92, $P<0.05$), ATRX maintained high balanced accuracy near 0.89 while AUC increased from 0.69 to 0.82, and Ki67 rose from 0.60 to 0.69 (AUC 0.50 to 0.51). For EGFR, TERT, and TP53, balanced accuracy increased (0.65 to 0.66; 0.81 to 0.85; 0.65 to 0.71, respectively) with mixed AUC changes. Variability generally decreased with +DRO; for example, CDKN2A/2B AUC error reduced from 0.44 to 0.12 and TP53 balanced-accuracy error from 0.04 to 0.006. Clinically, higher balanced accuracy indicates fewer missed positives and more reliable calls on uncommon markers, supporting more confident prognostic discussions, trial eligibility assessments, and treatment planning.\\
\begin{figure}
    \centering
    \includegraphics[width=1\linewidth]{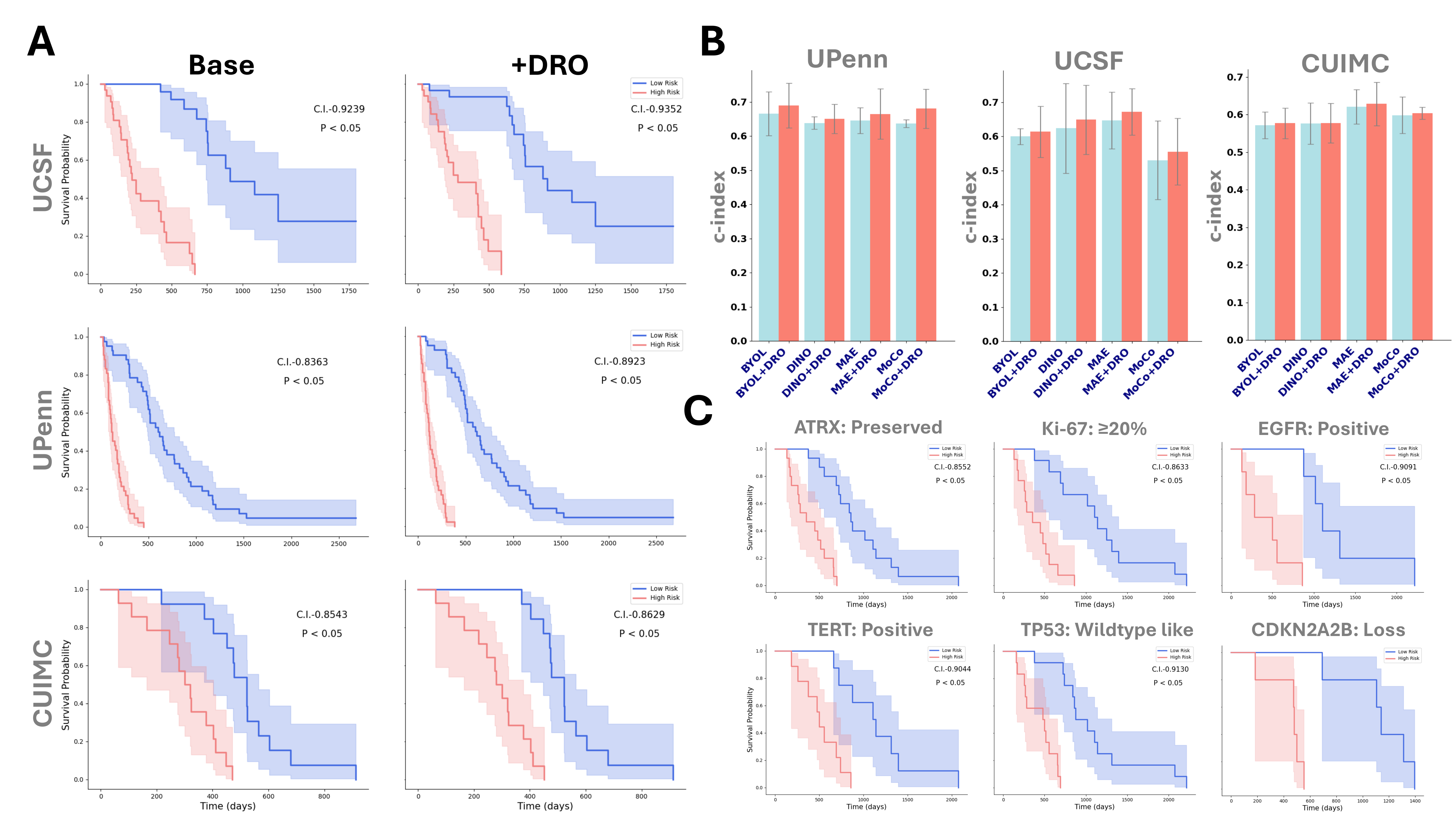}
    \caption{A. Comparison high- and low-risk patients of base SSL methods and +DRO methods for UCSF, UPenn and CUIMC cohorts, B. c-index for base and +DRO methods, C. Stratification of high- and low-risk patients for different Uncommon mutation types.}
    \label{fig:figure5}
\end{figure}
\noindent\textbf{Overall Survival Analysis:} We next assessed the prognostic value of foundation-model embeddings for overall survival in UCSF, UPenn, and CUIMC. Kaplan–Meier curves for IDH1 wild-type glioblastoma, stratified by model-derived risk score, are shown in Figure~\ref{fig:figure5}A. Averaged across BYOL, DINO, MAE, and MoCo encoders, the c-index increased with imbalance-aware training (DRO) at all three institutions: CUIMC rose from 0.592 to 0.597 ($P=0.0359$), UPenn from 0.600 to 0.623 ($P=0.0329$), and UCSF from 0.647 to 0.671 ($P=0.0041$). Figure~\ref{fig:figure5}B summarizes per-backbone results; DRO improved the mean c-index for every encoder across sites, with average gains of +0.014 (BYOL), +0.013 (DINO), +0.017 (MAE), and +0.025 (MoCo).
The strongest site-specific configurations were MAE+DRO at CUIMC (0.628 ± 0.058), MAE+DRO at UPenn (0.672 ± 0.068), and BYOL+DRO at UCSF (0.690 ± 0.065). Improvements at CUIMC were modest but consistent across backbones, while UPenn and UCSF showed clearer gains. Overall, these results indicate that DRO enhances the discriminative ability of survival models built on foundation-model embeddings and supports more reliable risk stratification across independent cohorts. In Figure~\ref{fig:figure5}C, Kaplan–Meier curves for the uncommon markers are shown stratified by model-derived risk score.\\
\begin{figure}
    \centering
    \includegraphics[width=0.8\linewidth]{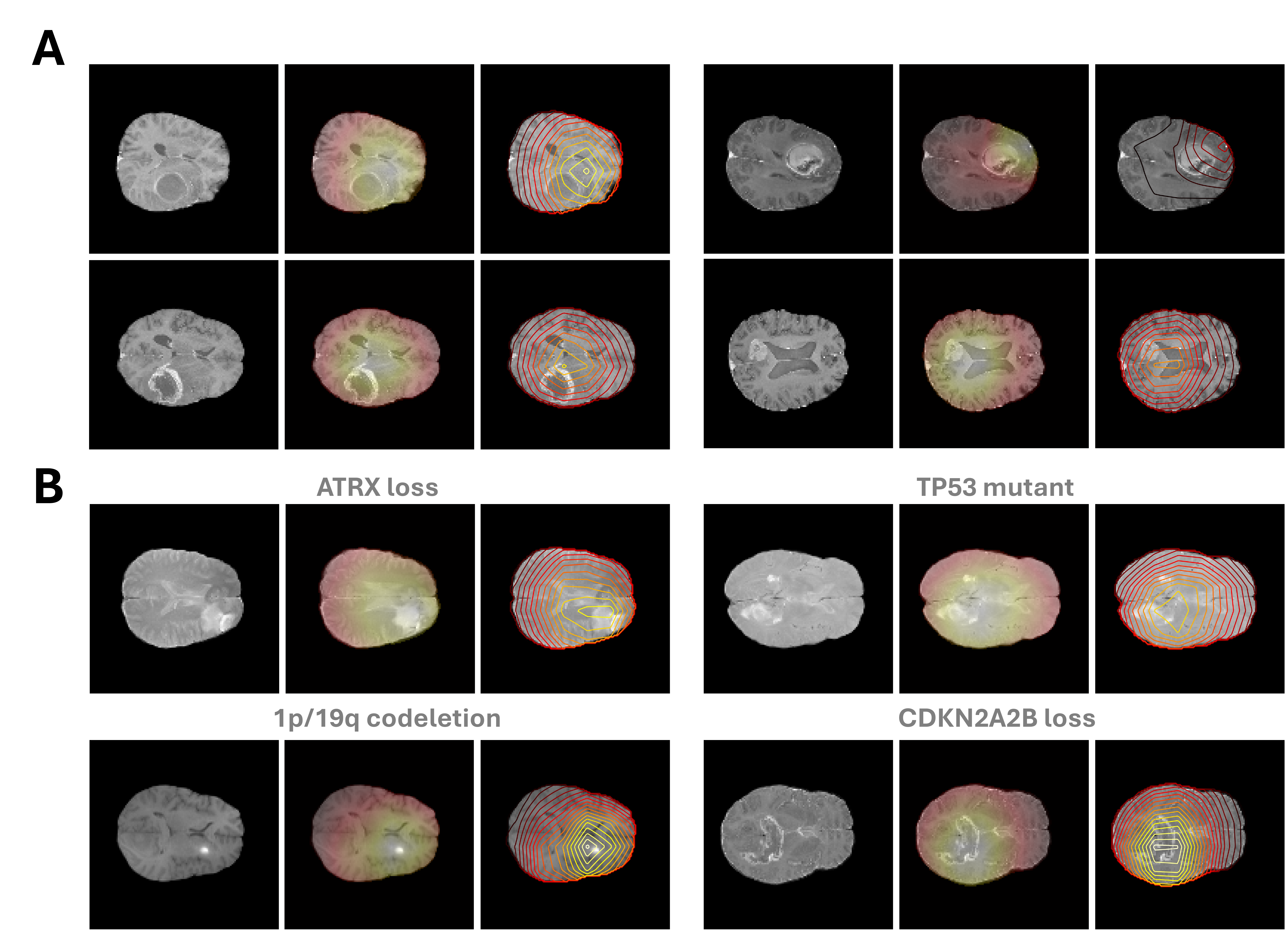}
    \caption{A. GradCAM heatmaps and contour maps for different cases, B. Visualizations of different uncommon mutation types.}
    \label{fig:figure6}
\end{figure}
\noindent\textbf{Model Interpretability:} To examine the biological and anatomical plausibility of model predictions, we performed interpretability analyses using Grad-CAM visualizations on the CUIMC dataset (Figure~\ref{fig:figure6}). Figure~\ref{fig:figure6}A presents representative cases, including two with MGMT unmethylated and IDH1 wild-type status, and three with MGMT methylated and IDH1 wild-type status. Figure~\ref{fig:figure6}B depicts Grad-CAM activation maps and prediction counters for uncommon molecular alterations, including ATRX loss, TP53 mutation, 1p/19q codeletion, and CDKN2A/2B loss. These visualizations reveal focused attention on tumor regions, providing qualitative evidence that the learned representations capture biologically meaningful and spatially relevant imaging features.
\section*{Discussion}
This work presents a segmentation-free, multi-sequence MRI foundation model tailored to neuro-oncology and demonstrates its utility for molecular profiling and prognostic risk stratification across three independent institutions (UCSF, UPenn, CUIMC). Embeddings produced by the model, when trained with distributionally robust optimization (DRO), consistently improved classification of common markers (e.g, MGMT, IDH1) and delivered measurable gains on uncommon or imbalanced endpoints, while reducing site-specific clustering in the representation space. For overall survival among IDH1 wild-type glioblastoma, DRO increased the c-index at each site, with the strongest site–backbone configurations reaching 0.628–0.690 and average gains observed across BYOL, DINO, MAE, and MoCo. These findings indicate that neuro-oncology–specific pretraining combined with imbalance-aware training yields representations that transfer across tasks and are more stable under real-world multi-center heterogeneity.
Clinically, improved balanced accuracy on molecular endpoints translates into fewer false negatives for under-represented alterations and more reliable triage when tissue is limited or delayed. The improved survival prediction suggests that our model captures biologically meaningful tumor signals rather than site-specific idiosyncrasies, supporting its use in guiding risk-adapted care, selecting patients for clinical trials, and planning follow-up imaging. Notably, the observed reductions in cohort separability (via PCA silhouette and t-SNE) together with consistent center-wise performance gains suggest that DRO down-weights spurious site signatures and reweights harder strata, yielding portability that is architectural-agnostic and complementary to diverse self-supervised backbones. By reducing the influence of scanner or institution differences, the approach offers a more reliable and broadly applicable tool that can integrate seamlessly into diverse clinical settings.

A key practical advantage of the approach is scalability without manual segmentation. Despite being segmentation-free, Grad-CAM visualizations concentrated on tumor and peri-tumoral regions for both common and uncommon markers, providing qualitative face validity of the learned features. This is attractive for clinical workflows where precise segmentations are costly. Nonetheless, attribution methods can be sensitivity-prone; future work should incorporate quantitative sanity checks (e.g., counterfactual occlusions) and alignment with expert annotations to strengthen interpretability claims. Hybrid objectives that weakly supervise masks could further bridge the gap between scalable pretraining and delineation-dependent tasks (e.g., volumetrics for response assessment).
Relative to classical radiomics pipelines that rely on handcrafted features and high-quality segmentations, the proposed framework learns tumor-specific features directly from raw multi-sequence MRI and scales across heterogeneous cohorts. Compared with purely supervised CNNs, foundation-model pretraining reduces annotation burden, and in conjunction with DRO, improves robustness to covariate shift. The consistent—albeit sometimes modest—site-wise c-index improvements and the broad gains on classification endpoints argue that performance benefits reflect better generalization rather than overfitting to any single dataset.

Several limitations merit careful attention. First, the analysis is retrospective, and thus vulnerable to unmeasured confounding. Differences in treatment timing and intensity—including extent of resection, radiation dose and field design, chemotherapy regimens, and the use of adjunctive steroids—may influence both imaging features and clinical outcomes, potentially biasing associations. Imaging schedules, which vary by institution and patient course, could also introduce heterogeneity in label assignment and survival estimation. Second, class imbalance and assay variability pose challenges for uncommon mutations and continuous markers such as Ki-67. Limited sample sizes for these endpoints increase the risk of label noise and reduce statistical power, raising uncertainty about the stability of performance gains. Third, while our approach improved discrimination, we did not fully characterize calibration or clinical utility. Future work should report calibration slope and intercept, Brier or integrated Brier scores (IBS), and decision curve analyses to ensure that predictions are both reliable and actionable in practice. Fourth, site separability metrics such as PCA silhouette and t-SNE are useful but ultimately proxy measures; more rigorous assessments of robustness are required, including vendor- and protocol-stratified analyses, evaluation across scanner generations, and targeted stress tests simulating distributional shift. Finally, survival modeling was restricted to IDH1 wild-type glioblastoma and did not account for time-varying covariates such as treatment changes, progression events, or recurrent therapies, which limits clinical granularity. Incorporating dynamic modeling approaches that integrate temporal, interventional, and multimodal data will be essential for building clinically deployable prognostic tools.

Future efforts will prioritize prospective, pre-registered validation across diverse geographic regions and healthcare systems to ensure reproducibility and generalizability beyond academic centers. Such studies will incorporate comprehensive calibration analyses and decision-curve–based utility assessments to establish that model predictions are not only accurate but also clinically actionable. In parallel, integrating explicit temporal and interventional modeling—such as dynamic survival heads, recurrent event prediction, and treatment-conditioned objectives—will enable the framework to better capture disease trajectories and simulate “what-if” scenarios within digital-twin frameworks. These capabilities could support individualized treatment planning, adaptive trial design, and longitudinal risk monitoring. Expanding the foundation model to multimodal fusion with genomic, pathologic, and clinical data will further contextualize imaging-derived signals, while principled approaches to uncertainty quantification (e.g., Bayesian inference, conformal risk control) will be critical for communicating prediction confidence in high-stakes settings. Finally, systematic fairness audits across demographic, socioeconomic, and institutional strata are essential to ensure equitable model performance and to prevent exacerbation of healthcare disparities, particularly as these tools move closer to real-world deployment.

\section*{Conclusion}
Our proposed neuro-oncology–specific foundation model coupled with DRO produces more site-invariant, task-transferable representations that improve molecular marker prediction—including on uncommon markers—and stabilize prognostic discrimination across institutions. With prospective validation, stronger calibration, and longitudinal/ interventional extensions, such models have the potential to operationalize precision neuro-oncology by accelerating molecular triage, informing risk-adapted care, and enabling robust stratification in clinical trials.

% \section*{Introduction}
% % Introduce the problem, its significance, and provide background.

% \section*{Materials and Methods}
% \textbf{Study Design and Overview.}\\
% \textbf{Study Population.}\\
% \textbf{Training.}\\
% \textbf{Statistical Analysis.}\\

% \section*{Results}
% % Present the key findings, including tables and figures.

% \section*{Discussion}
% % Interpret the results, compare with previous studies, and discuss implications or limitations.

\bibliographystyle{unsrt}
\bibliography{main}

\clearpage

\clearpage
\clearpage
\clearpage

\clearpage

\clearpage

% \clearpage
% \input{tables/table_3}

\clearpage
\section*{Appendix}
\begin{table}[htbp]
\centering
\caption{Supplementary Table (Training): Dataset Composition and Sample Size}
\begin{tabular}{|l|c|}
\hline
\textbf{Dataset} & \textbf{N} \\
\hline
BraTS-GLI-Pre      & 1443 \\
BraTS-GLI-Post     & 1809 \\
BraTS-MEN          & 1141 \\
BraTS-MET          & 1475 \\
BraTS-PED          & 351  \\
LUMIERE            & 599  \\
MU-GLioma-Post     & 596  \\
% CUMC               & --   \\
\hline
\textbf{Total}     & 7414 (--) \\
\hline
\end{tabular}
\label{tab:dataset_summary}
\end{table}

\clearpage% \begin{table}[htbp]
% \centering
% \caption{Supplementary Table: Sample Sizes by Dataset and Institution}
% \begin{tabular}{|l|l|c|}
% \hline
% \textbf{Dataset} & \textbf{Institution} & \textbf{N} \\
% \hline
% \multirow{3}{*}{AG}  & UCSF   & 111 \\
%                      & UPENN  & 95  \\
%                      & CUMC   & 173 \\
% \hline
% \multirow{2}{*}{MET} & CUMC   & --  \\
%                      & UCSF   & 137 \\
% \hline
% MEN                  & USC    & 96  \\
% \hline
% \end{tabular}
% \label{tab:site_dataset_summary}
% \end{table}

\begin{table}[htbp]
\centering
\caption{Supplementary Table: Sample Sizes by Dataset and Institution}
\begin{tabular}{|l|c|}
\hline
\textbf{Institution} & \textbf{N} \\
\hline
UCSF   & 111 \\
UPENN  & 95  \\
CUIMC   & 292 \\
% \multirow{3}{*}{AG}  & UCSF   & 111 \\
%                      & UPENN  & 95  \\
%                      & CUMC   & 173 \\
% \hline
% \multirow{2}{*}{MET} & CUMC   & --  \\
%                      & UCSF   & 137 \\
% \hline
% MEN                  & USC    & 96  \\
\hline
\end{tabular}
\label{tab:site_dataset_summary}
\end{table}

\end{document}